\documentclass[sigconf]{acmart}

\AtBeginDocument{%
  }

\copyrightyear{2023}
\acmYear{2023}
\setcopyright{acmlicensed}\acmConference[MM '23]{Proceedings of the 31st ACM International Conference on Multimedia}{October 29-November 3, 2023}{Ottawa, ON, Canada}
\acmBooktitle{Proceedings of the 31st ACM International Conference on Multimedia (MM '23), October 29-November 3, 2023, Ottawa, ON, Canada}
\acmPrice{15.00}
\acmDOI{10.1145/3581783.3613809}
\acmISBN{979-8-4007-0108-5/23/10}

\usepackage{balance}
\usepackage{subfigure}
\usepackage{multirow}

\begin{document}

\title{Tile Classification Based Viewport Prediction with \\ Multi-modal Fusion Transformer}


\author{Zhihao Zhang}
\orcid{0000-0002-3791-8785}
\authornote{Both authors contributed equally to this research.}
\affiliation{%
  \institution{Xi'an Jiaotong University  }
  \streetaddress{}
  \city{Xi'an}
  \country{China}}
\email{zh1142@stu.xjtu.edu.cn}
\author{Yiwei Chen}
\authornotemark[1]
\orcid{0000-0002-0219-9143}
\affiliation{%
  \institution{Xi'an Jiaotong University }
  \streetaddress{}
  \city{Xi'an}
  \country{China}}
\email{chenyiwei2000@stu.xjtu.edu.com}

\author{Weizhan Zhang}
\orcid{0000-0003-0330-5435}
\authornote{Corresponding author.}
\affiliation{%
  \institution{Xi'an Jiaotong University }
  \streetaddress{}
  \city{Xi'an}
  \country{China}}
\email{zhangwzh@mail.xjtu.edu.cn}

\author{Caixia Yan}
\orcid{0000-0001-7763-2987}
\affiliation{%
  \institution{Xi'an Jiaotong University }
  \streetaddress{}
  \city{Xi'an}
  \country{China}}
\email{yancaixia@xjtu.edu.cn}

\author{Qinghua Zheng}
\orcid{0000-0002-8436-4754}
\affiliation{%
  \institution{Xi'an Jiaotong University }
  \streetaddress{}
  \city{Xi'an}
  \country{China}}
\email{qhzheng@mail.xjtu.edu.cn}

\author{Qi Wang}
\orcid{0000-0002-0639-2119}
\affiliation{%
  \institution{MIGU Video}
  \streetaddress{}
  \city{Shanghai}
  \country{China}}
\email{wangqi@migu.cn}

\author{Wangdu Chen}
\orcid{0000-0002-8921-0092}
\affiliation{%
  \institution{MIGU Video}
  \streetaddress{}
  \city{Shanghai}
  \country{China}}
\email{chenwangdu@migu.cn}

\renewcommand{\shortauthors}{ Zhihao Zhang et al.}


\begin{abstract}
Viewport prediction is a crucial aspect of tile-based $360^{\circ}$ video streaming system. 
However, existing trajectory based methods lack of robustness, also oversimplify the process of information construction and fusion between different modality inputs, leading to the error accumulation problem. 
In this paper, we propose a tile classification based viewport prediction method with Multi-modal Fusion Transformer, namely MFTR.
Specifically, MFTR utilizes transformer-based networks to extract the long-range dependencies within each modality, then mine intra- and inter-modality relations to capture the combined impact of user historical inputs and video contents on future viewport selection. 
In addition, MFTR categorizes future tiles into two categories: user interested or not, and selects future viewport as the region that contains most user interested tiles.
Comparing with predicting head trajectories, choosing future viewport based on tile's binary classification results exhibits better robustness and interpretability.
To evaluate our proposed MFTR, we conduct extensive experiments on two widely used PVS-HM and Xu-Gaze dataset. 
MFTR shows superior performance over state-of-the-art methods in terms of average prediction accuracy and overlap ratio, also presents competitive computation efficiency.
\end{abstract}

\ccsdesc[300]{Computing methodologies~Planning under uncertainty}
\ccsdesc[500]{Information systems~Multimedia information systems}

\keywords{viewport prediction; multi-modal fusion; transformer network; tile classification}



\maketitle

\section{Introduction}
With the rapid growth of Virtual Reality(VR), $360^{\circ}$ video has gained more attention due to its immersive and interactive experience to viewers. 
There are tons of $360^{\circ}$ videos on major video platforms such as YouTube and Meta. 
However, streaming $360^{\circ}$ video demands massive bandwidth compared to transfer regular planar video \cite{yang2019single}. 
For example, transmitting a 4K $360^{\circ}$ video requires a network throughput of $25$ Mbps and above while a 4K regular video only needs $5-10$ Mbps. 
Thus, $360^{\circ}$ video delivery under limited bandwidth network is challenging and needs to be optimized.

Recently, tile-based streaming approaches \cite{xie2017360probdash, van2019optimizing, 
 nguyen2018your, chen2020sparkle} have been proposed, they divide each $360^{\circ}$ video frame into equal sized tiles, and stream users' viewport which consists of several contiguous tiles in higher resolution. 
Specifically, prevailing methods forecast users' head position coordinates first, and then select a viewport whose geometric center is closest to the coordinates.
\begin{figure}[t]
 \setlength{\abovecaptionskip}{0.cm}
  \centering
    \subfigure[Trajectory based]{              
        \includegraphics[width=0.45\linewidth]{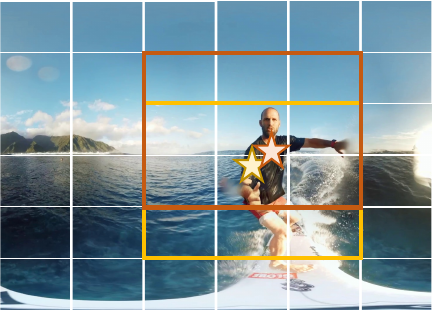}
        \label{fig1a}}
    \subfigure[Tile classification based]{
        \includegraphics[width=0.45\linewidth]{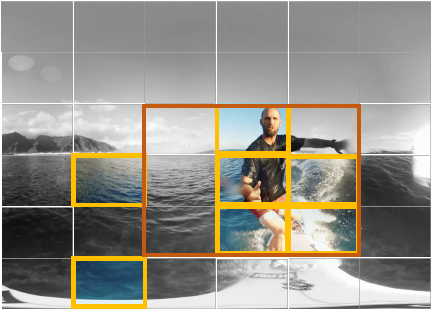}
        \label{fig1b}}
        \caption{ (a) With a slight difference of predicted head coordinates (colored stars), selected future viewports (colored bounding boxes) exhibit noticeable positional deviations. (b) The proposed MFTR method based on tile classification, which divides tiles into two two categories: user-interested tiles (colored) and non-user interested tiles (colorless). The predicted result (red bounding box) is a region of viewport-size that contains the majority of user-interested tiles.}
\vspace{-0.5cm}
\end{figure}
For example, \cite{de2019delay,yang2019single,rondon2019revisiting,van2022machine} concentrate on temporal dimension, encode historical temporal features and forecast head positions, transfering viewport prediction task to typical time series prediction problem.
However, these approaches ignore the importance of visual contents on users' focus changes, and predict head positions in a step-by-step way which leads to an error accumulation problem. 
To alleviate these limitations, methods such as \cite{aladagli2017predicting,xu2018gaze,feng2019viewport,zhang2022mfvp,li2022spherical} take both visual and temporal features into consideration, utilize CNN based networks along with recurrent neural network (RNN) to predict the coordinates of head positions jointly. 
Although noticeable improvements have been made by fusing two modality features to forecast viewport jointly, these approaches simplify the process of information construction and fusion, while the prediction process is still mechanized.

To be more specific, there are two main defects in previous approaches. 
First, prevailing methods are both trajectory based, forecast viewport through predicting head positions, which lack of robustness. 
For example, as shown in Fig.\ref{fig1a}, with a small shift in the predicted head coordinates, the corresponding viewports are obviously different.  
Therefore, a more stable viewport prediction mechanism should be designed.
Second, these conventional methods simplify feature construction process and fuse different modality features using mechanized processes like concatenation, complete the prediction task in a step-to-step way. 
These network structures lead to an error accumulation problem, resulting in a drastic drop in prediction accuracy with the prolongation of the prediction time. 
Thus, we argue that a more comprehensive network structure should be considered.

In order to solve the problem of low robustness, we propose a tile-classified scheme, predict the final viewport based on tile binary classification results rather than head trajectary coordinates. 
As illustrated in Fig.\ref{fig1b}, we classify all the tiles into two categories, one is the tiles that users are interested in, and the others are not, and then choose the viewport with highest overlapping ratio with the interested tiles. 
By doing so, proposed tile classification based approach selects the viewport with the highest probability of user interest instead of single head position, which is more robust and interpretable.

Benefiting from the success of transformer architecture \cite{vaswani2017transformer} in modality relationship modeling, we propose a transformer-based prediction framework, which explicitly considers both visual and temporal features to jointly forecast the viewport of interest. 
Our method employs multi-head attention module along with learnable tokens to effectively mine interactions between modalities and accurately classify the interested tiles.
In this sense, we call it Multi-modal Fusion Transformer(MFTR). 
The proposed MFTR includes five key designs.
First, Temporal Branch contains two LSTM networks to encode head and eye movements respectively, followed by a Temporal Transformer to mine the motion trend within the same modal and extract the long-range temporal dependencies. 
Second, Visual Branch adopts MobileNetV2 backbone along with a Visual Transformer to obtain more robust visual representations.
Then, we add extra modality token on temporal and visual embeddings to represent different modal properties, after which Temporal-Visual Fusion Module utilizes Temporal-Visual Transformer to explore intra- and inter-modality relations and aggregate them.
Moreover, Position Prediction Head supports the training of Temporal Transformer by providing more supervision information about fused temporal features.
At last, in Tile Classification Head, we generate score map for each tile in every time stamp to represent the probability that each tile may be concerned, and the final viewport is the region which contains most tiles with the score above the threshold.

To summary, the major contributions are as follows: 
\begin{itemize}
\item We propose a novel tile classification based viewport prediction scheme, transfer viewport prediction to a binary tile classification task providing better robustness and interpretability.
\item We propose a Multi-modal Fusion Transformer network namely MFTR, extracting the long-range dependencies within single modality and modeling the combined influence of both temporal and visual features on the tiles that users interested in.
\item We demonstrate our method outperforms state-of-the-art approaches on two widely used datasets, shows higher prediction accuracy with competitive computation efficiency especially for long term prediction.
\end{itemize}

\section{Related Work}
\subsection{Viewport Prediction}
Viewport prediction task \cite{yaqoob2020survey,chiariotti2021survey} 
aims to predict the future viewport based on user's past viewing history. 
There are basically two kinds of schemes to predict future viewport: 
\textit{Content-Agnostic Approaches}: 
\cite{de2019delay,yang2019single,rondon2019revisiting,jamali2020lstm,nasrabadi2020viewport,chopra2021parima,lee2021prediction,van2022machine} 
leverage typical networks to capture temporal dependencies in order to predict future head motions. 
\cite{jamali2020lstm} took user’s viewpoint position history as sequences and predict the future viewpoint positions through an LSTM network. 
\cite{chao2021transformer} leveraged the past viewport scan path to predict user's future directions of movements. \textit{Content-Aware Approaches}: 
Unlike Content-Agnostic Approaches, \cite{aladagli2017predicting, xu2018gaze,ban2018cub360, zhu2019prediction, feng2019viewport, xu2019analyzing, li2019very, jiang2020svp, yaqoob2021combined, jiang2022robust, zhang2022vrformer} predict future viewport using temporal features and visual contents jointly.
\cite{zhang2022mfvp} utilizes a graph convolutional network (GCN) to output the saliency maps of the 360 video frames.
Moreover, \cite{li2022spherical} propose a spherical convolution-empowered viewport prediction method to eliminate the projection distortion of 360-degree video. 

MFTR method belongs to Content-Aware Approaches, but different from the above trajectory based methods aim at predicting future head motions, we transfer the viewport prediction to a tile classification task, providing a more robust and stable method which strives to find out tiles of user interest. 

\begin{figure*}[ht]
\setlength{\abovecaptionskip}{0.cm}
  \centering
    \subfigure[Encoder Layer]{              
        \includegraphics[width=0.1885\linewidth]{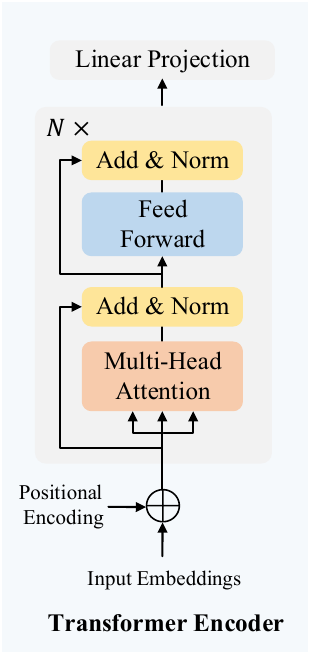}
        \label{modela}}
    \subfigure[MFTR Model Overview]{
        \includegraphics[width=0.75\linewidth]{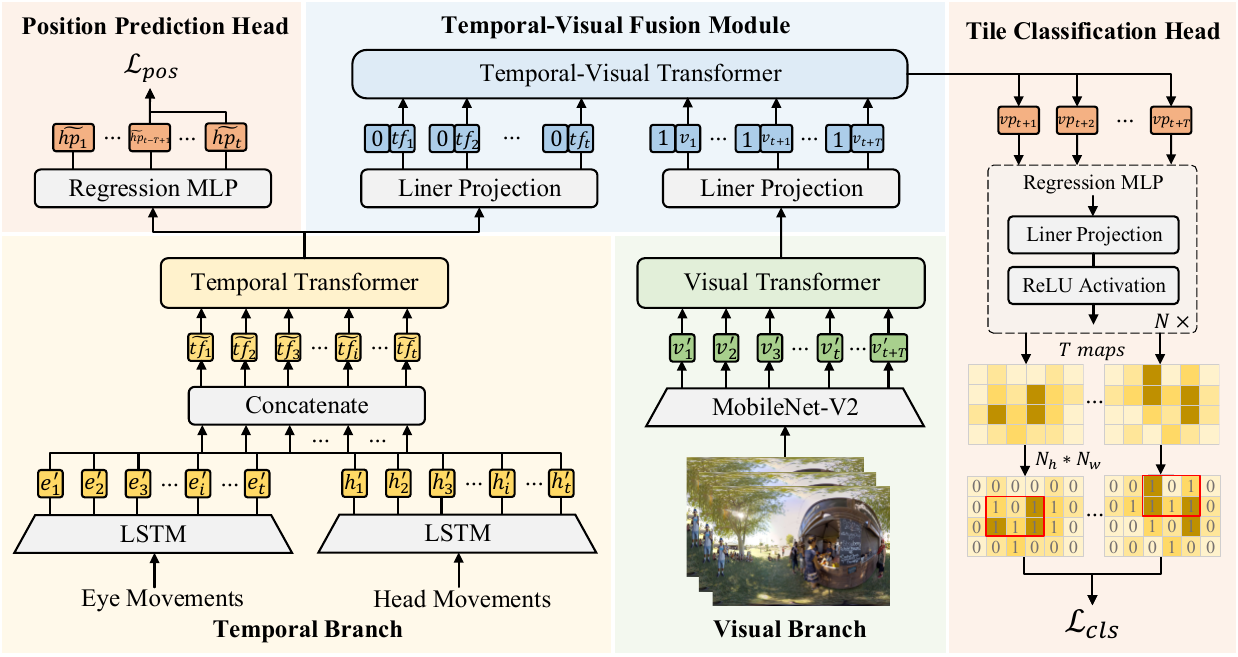}
        \label{modelb}}
    \caption{(a) Transformer encoder architecture revisit. (b) An overview of our proposed framework. MFTR contains five components: (1) Temporal Branch extracts fused temporal feature from head, eye movement histories; (2) Visual Branch encodes the visual features from given 360 video frames, and enhances them by Visual Transformer; (3) Temporal-Visual Fusion Module fuses different modal features to indicate user's interested areas influenced by motion trend and visual area; (4) Position Prediction Head provides more supervision information and assist in the training of Temporal Transformer; (5) Tile Classification Head determines the user interested tiles and choose the future viewport.}
\end{figure*}

\subsection{Transformer in Multi-modal Tasks}
Transformer is first proposed in \cite{vaswani2017transformer} to handle neural machine translation task. 
Attention module, as the core part in transform, focus on aggregating the information of the whole sequence with adaptive weights, could fully capture long-term dependency in parallel. 
Recently, transformer networks have achieved great success in NLP field \cite{devlin2018bert,kang2020natural, raffel2019t2t} and vision tasks \cite{dosovitskiy2020image,carion2020end, liu2022video, han2022survey, liu2021swin}. 
Inspired by transformer's intrinsic advantages in modeling different modalities, a large number of transformers have been studied extensively for various multi-modal tasks.
Visual-linguistic pre-training \cite{kim2021vilt,lu2019vilbert,lu202012,li2020unicoder} construct representations of images and texts jointly. 
In general, multi-modal transformer works need to perform two steps: tokenize the input and devise several transformer encoder layers for joint learning. 
In our work, we perform transformer capturing long-range dependencies within single modaity and modeling the combined influence of multi-modal features through multi-modal transformers. 

\section{Method}
\subsection{Problem Formulation}
Viewport prediction is to predict user's future viewport for the incoming seconds conditioned on user historical traces and $360^{\circ}$ video contents. 
Formally, given temporal inputs, include past head orientations $\tilde{H}=\left\{\tilde{h}_{1}, ..., \tilde{h}_{t} \right\}$ and eye movements $\tilde{E}=\left\{\tilde{e}_{1}, ..., \tilde{e}_{t} \right\}$ from time stamp $1$ to $t$.
Where $\tilde{h}_{i}=\left\{(\phi_{i}, \varphi_{i})|-\pi\leq\phi_{i}\le\pi, -\pi/2\leq\varphi_{i}\le\pi/2\right\}$ and $\tilde{e}_{i}= \left\{(x_{i}, y_{i}) | 0\leq x_{i}\le 1,0\leq y_{i}\le1\right\}$ denotes the polar coordinates and the cartesian coordinates at time stamp $i$ respectively. 
As visual inputs, $\tilde{V} = \left\{\tilde{v}_{1}, ..., \tilde{v}_{t}, ..., \tilde{v}_{t+T} \right\}$ express sequences of $360^{\circ}$ video frames where $\tilde{v}_{i}$ corresponds to visual contents at time $i$. 
Moreover, $\left\{t+1,...,t+T\right\}$ are defined as predicted time window, $T$ is the prediction length. 
The overall goal is to predict user's viewport in the predicted time window, formulated as $P=\left\{p_{t+1}, ..., p_{t+T} \right\}$, where $p_{t+i}$ means the position coordinates of viewport. We need to train a objective function $f(\cdot)$ with learnable parameters $W^*$ given temporal and visual inputs,
\begin{equation}
\begin{aligned}
    P = f \left(\tilde{H}, \tilde{E}, \tilde{V}; W^{*} \right).
\end{aligned} 
\end{equation}
\subsection{Multi-head Attention Revisit}
Before detailing the architecture of MFTR, we first revisit the Multi-Head Attention mechanism, which is the core module of our framework. Given query $Q$, key $K$, and value $V$ with $d$ channel dimensions, Multi-Head Attention \cite{vaswani2017transformer} calculates the attended output as:
\begin{equation}
\begin{gathered}
    Att(Q, K, V)=Softmax({QK^T}/{\sqrt{d_k}})V,  \\
    h_i=Att(QW_i^Q, KW_i^K, VW_i^V),  \\
    MultiHead(Q, K, V)=Concat(h_1,\cdots,h_n)W^O, \\
\end{gathered}
\end{equation}
where the parameter matrices $W_i^Q, W_i^K\in\mathbb{R}^{d \times d_k}$, $W_i^V\in\mathbb{R}^{d \times d_v}$, and $W^O\in\mathbb{R}^{nd_v \times d}$, and $n$ represents the number of parallel attention layers. 
Specifically, when setting  $Q$, $K$, $V$ to the same value, the module called Multi-head Self-Attention (MSA). 

The typical transformer has encoder-decoder structure, while we only use transformer encoder as the basic block, like Figure \ref{modela} illustrated. Each transformer encoder has two sub-networks: MSA and feed forward network (FFN), and designed in residual structures. Concretely, let us denote the input as $x_i$, the transformer encoder's output is:
\begin{equation}
\begin{gathered}
    x_i^{\prime}=f_{LN}(x_i+f_{MSA}(x_i)),  \\
    x_{i+1}=f_{LN}(x_i^{\prime}+f_{FFN}(x_i^{\prime})),
\end{gathered}
\end{equation}
where $f_{LN}$ means layer normalization, $f_{MSA}$ indicates multi-head self-attention outputs, and $f_{FFN}$ implies feed-forward network which is composed of fully connected layers and ReLU activation layers. 

\subsection{Multi-modal Fusion Transformer}

We propose a tile classification based multi-modal transformer (MFTR) for viewport prediction task which is depicted in Fig. \ref{modelb}.
The model consists of five main building blocks: 
(1) \textit{Temporal Branch} utilizes Temporal Transformer to construct comprehensive temporal features from past head and eye movements; 
(2) \textit{Visual Branch} generates robust visual features from given $360^\circ$ video frames; 
(3) \textit{Temporal-Visual Fusion Module} apply transformer to capture intra- and inter-modality contexts; 
(4)\textit{Position Prediction Head} forecasts head position as supervision information during training phase; 
(5) \textit{Tile Selection Head} chooses user interested tiles and forecast the future viewport. 
In the following, we explain each module of MFTR in details.

\paragraph{Temporal Branch.}
Different from time series problems, viewport switching is more objective which means there are many user-side factors that affect their choice for the future viewport. We posit that user historical head movements and eye motions have a decisive role in future viewport choosing.
Based on this, to better indicate users' moving habits, we employ two long shot term memory (LSTM) networks to encode temporal inputs respectively, and use Temporal Transformer to fuse them. 
Both two LSTMs have hidden size 256, which means the representations of head and gaze positions share same channel dimension.  
The Temporal Transformer composed by 6 transformer encoders \cite{vaswani2017transformer}, and the channel dimension set as 512.

Given the past head movements $\tilde{H}\in \mathbb{R}^{t\times 2}$ and eye histories $\tilde{E} \in \mathbb{R}^{t\times 2}$, after leveraging fully connected layer (FC) to increase channel dimensions, we obtain $H \in \mathbb{R}^{t\times C_h} $, $E \in \mathbb{R}^{t\times C_e}$.
Then we employ two LSTM networks to generate temporal embeddings $H^{\prime}$ and $E^{\prime}$, which share the same shape as $H$, $E$.  
Through concatenation, the temporal features $\widetilde{TF} \in \mathbb{R}^{t\times C_{tf}}$ can be acquired, where $C_{tf}=C_h+C_e=512$.
After that, we use a Temporal Transformer to mine the motion trend within single modality, and gain the fused temporal features $TF =\left\{tf_1,tf_2,\cdots,tf_t\right\} \in \mathbb{R}^{t\times C_{tf}}$.

\paragraph{Visual Branch.} 
$360^\circ$ video content is another a key factor since users can also be attracted by the contents in the videos.
In Visual Branch, we start with a backbone network and followed by Visual Transformer to acquire more robust visual representions. 
Unlike typical convolutional backbones, we employ MobileNet-V2\cite{sandler2018mobilenetv2} because of its lightweight and portability. 
Concretely, MobileNet-V2 generates visual features of 1000 channel dimensions for each video frame.
And Visual Transformer shares the same structure with Temporal Transformer.

Specifically, given the input frames $\tilde{V}\in\mathbb{R}^{(t+T) \times 3 \times H \times W}$, we exploit MobileNet-V2 network to generate visual features $V^{\prime} \in\mathbb{R}^{(t+T) \times 1000}$.  
Then, we leverage a fully connected layer to reduce the dimension and obtain newly $V^{\prime\prime}\in\mathbb{R}^{(t+T) \times C_v}$, where we set $C_v$ as 512.
Moreover, we further add position embeddings on $V^{\prime\prime}$ \cite{carion2020end} to let the transformer encoder sensitive to positions of visual sequence. 
Visual Transformer explores the sequence of image features and output more robust representations  $V=\left\{v_1,\cdots,v_{t+1},\cdots,v_{t+T}\right\}\in\mathbb{R}^{(t+T) \times C_v}$ which share the same size with $V^{\prime\prime}$. 

\paragraph{Temporal-Visual Fusion Module.} 
Thanks to the attention mechanism for its good performance in modality relationship modeling.
Temporal-Visual Fusion Module is also a stack of $6$ transformer encoders, 
to explore intra- and inter-modality relations and aggregate temporal-visual features together. 
Specifically, since the fused temporal embeddings $T$ and visual representations $V$ share the same channel dimension, the joint input for this fusion module can be formulated as:
\begin{equation}
\begin{aligned}
    In  = \overbrace{ [tf_{1}+tf_{2}+\cdots+tf_{t},}^{temporal features}\underbrace{v_{1},\cdots,v_{t},\cdots,v_{t+T}]}_{visual features},
\end{aligned} 
\end{equation}
where each $tf_i$ and $v_i$ has same dimension $C=C_{tf}=C_v=512$.
Following \cite{kim2021vilt} we further add a modal-type embedding on the joint input to let the model distinguish the differences between two modalities.
Because fused results represent the combined impact of temporal and visual features while $\left\{v_{t+1},\cdots,v_{t+T}\right\}$ imply the visual contents in predicted time window, we apply the transformer outputs of these features $VP=\left\{vp_{t+1},\cdots,vp_{t+T}\right\} \in \mathbb{R}^{T \times C}$ for the next stage of prediction.

\paragraph{Position Prediction Head.} 
During training phase, we forecast the possible head positions in the predicted time window to provide more supervision information and assist in the training of Temporal Transformer. 
Since Temporal Transformer could mine long-distance dependencies in temporal dimension, we employ a basic MLP with hidden size 128 and 64 to forecast the possible head motions. 
More Concretely, MLP regresses the channel dimension $C_{tf}$ of temporal features $TF$ into 2 and output $\widetilde{HP}=\left\{\widetilde{hp}_1,\cdots,\widetilde{hp}_{t-T+1},\cdots,\widetilde{hp}_t\right\} \in\mathbb{R}^{t \times 2}$.
We select last $T$ elements $HP=\left\{\widetilde{hp}_{t-T+1},\cdots,\widetilde{hp}_t\right\}=\left\{hp_1,\cdots,hp_T\right\}\in\mathbb{R}^{T \times 2}$ as the predicted head movement results. 
We adopt Mean-Squared Loss to measure the quality of temporal feature fusion, 
thus the regression loss $\mathcal{L}_{pos}$ for Position Prediction Head is defined as:
\begin{equation}
\begin{aligned}
    \mathcal{L}_{pos}  = \frac{1}{T} \sum_{i=1}^T(hp_i-\hat{hp}_i)^2,
\end{aligned} 
\end{equation}
where $hp_i$ is predicted head movements while $\hat{hp}_{i}$ represents the ground truth head coordinates at time stamp $i$. 

\paragraph{Tile Classification Head.} 
Since trajectory based approaches lack of robustness, we predict future viewport based on tile classification results. 
Every frame could be divided into $N=N_h \times N_w=10\times 20$ tiles in our work, so that we apply a simple MLP on transformer fused feature $vp_i$ to generate score map $S^i\in\mathbb{R}^{N_h\times N_w}$ at time stamp $i$. 
Each element $S_{m,n}^i \in [0,1]$ indicates how interested the user is in each tile, then we set a score threshold $\gamma = 0.55$ to classify the tiles into two categories: the tiles whose scores exceed $\gamma$ represent user-interested while the others are not. 
Score map can be turned to a 0-1 matrix $Y^i\in\mathbb{R}^{N_h\times N_w}$.
Then we select viewport-sized regions $P=\left\{p_{t+1},\cdots, p_i,\cdots ,p_{t+T}\right\}$  which contain most user interested tiles as the predicted viewports for each time stamp $i$. 
\begin{table*}[t]
    \centering
    \resizebox{.75\linewidth}{!}{
    \begin{tabular}{c|c|c|c|c|c|c|c|c|c|c}
    \toprule  
    Prediction length& \multicolumn{2}{c|}{1s}& \multicolumn{2}{c|}{2s}& \multicolumn{2}{c|}{3s}& \multicolumn{2}{c|}{4s}& \multicolumn{2}{c}{5s}\\
    \midrule
    Metrics& AP& AO&  AP& AO&  AP& AO&  AP& AO&  AP& AO\\
    \midrule  
    Offline-DHP& 0.787& 0.833& 0.667& 0.746&  0.542& 0.598& 0.465& 0.531& 0.409 &0.447\\
    VPT360& 0.850& 0.904& 0.745& 0.821&  0.658& 0.713&  0.604& 0.641&  0.559 &0.605\\
    SPVP360&  0.901& 0.937& 0.820& 0.887&  0.756& 0.811&  0.684& 0.728&  0.602 &0.650\\
    MFTR& \textbf{0.954}& \textbf{0.981}&  \textbf{0.872}& \textbf{0.905}& \textbf{0.825}& \textbf{0.865}&  \textbf{0.773}& \textbf{0.814}& \textbf{0.730} &\textbf{0.787}\\
    \bottomrule 
    \end{tabular}}
    \caption{Average prediction accuracy (AP) and average overlap ratio (AO) comparison with baseline methods on PVS-HM dataset.}
    \label{fov_accuracy_table}
    \vspace{-0.5cm}
\end{table*}

\begin{table*}[t]
    \centering
    \resizebox{.75\linewidth}{!}{
    \begin{tabular}{c|c|c|c|c|c|c|c|c|c|c}
    \toprule  
    Prediction length& \multicolumn{2}{c|}{1s}& \multicolumn{2}{c|}{2s}& \multicolumn{2}{c|}{3s}& \multicolumn{2}{c|}{4s}& \multicolumn{2}{c}{5s}\\
    \midrule
    Metrics& AP& AO&  AP& AO&  AP& AO&  AP& AO&  AP& AO\\
    \midrule  
    Offline-DHP& 0.694& 0.754& 0.636& 0.645&  0.559& 0.589& 0.511& 0.537& 0.434 &0.491\\
    VPT360& 0.747& 0.813& 0.662& 0.726&  0.610& 0.652&  0.557& 0.594&  0.486 &0.533\\
    SPVP360&  0.824& 0.857& 0.731& 0.783&  0.639& 0.695&  0.573& 0.630&  0.518 &0.566\\
    MFTR& \textbf{0.856}& \textbf{0.905}&  \textbf{0.767}& \textbf{0.819}& \textbf{0.702}& \textbf{0.754}&  \textbf{0.651}& \textbf{0.698}& \textbf{0.597} &\textbf{0.655}\\
    \bottomrule 
    \end{tabular}}
    \caption{Average prediction accuracy (AP) and average overlap ratio (AO) comparison with baseline methods on Xu-Gaze dataset }
    \label{fov_accuracy_table_xu}
    \vspace{-0.5cm}
\end{table*}
Since we have done binary classification task of all tiles, we adopt basic Cross-Entropy Loss as the loss function in this module, like:
\begin{equation}
\begin{split}
    \mathcal{L}_{cls}  =&  - \frac{1}{Num}\sum_{i=t+1}^{t+T} \sum_{m=1}^{N_w} \sum_{n=1}^{N_h} ((\hat{Y}_{m,n}^i \cdot log(S_{m,n}^i) \\
    &+ (1-\hat{Y}_{m,n}^i)\cdot log(1-S_{m,n}^i)),
\end{split} 
\end{equation}
where $Num=T\times N_w \times N_h$, and $\hat{Y}^i\in\mathbb{R}^{N_h \times N_w}$ is a 0-1 matrix which could represent whether tile is included in the ground truth viewport positions $\hat{p_i}$, each element in the matrix $\hat{Y}_{m,n}^i\in \left\{0,1\right\}$.

\subsection{Training Objective}
The above MFTR formulation yeilds a fully supervised end-to-end training with a total objective:
\begin{equation}
\label{loss}
\begin{aligned}
    \mathcal{L}  = \alpha \mathcal{L}_{pos} + \beta \mathcal{L}_{cls},
\end{aligned} 
\end{equation}
where $\alpha$ and $\beta$ are hyper parameters to balance these two losses, $\mathcal{L}_{pos}$ and $\mathcal{L}_{cls}$ are the loss function of Position Prediction Head and Tile Classification Head respectively.

\section{Experiments}
In this section, we evaluate proposed method MFTR and analyze the results.

\subsection{Experimental Setting}
\paragraph{Datasets.} 
Our experiments are conducted on two widely-used datasets: PVS-HM \cite{xu2018predicting} dataset and Xu-Gaze \cite{xu2018gaze} dataset. 
PVS-HM dataset comprises $76$ $360^\circ$videos with resolutions ranging from $3K$ to $8K$. 
Total $58$ participants viewed all the videos, used an HTC Vive to capture their head movements and eye fixation traces at the frame level.
In addition, Xu-Gaze dataset consists of 208 high definition dynamic $360^\circ$ videos with a frame rate of $25$. 
Total $45$ participants watched these videos, more than 900,000 head movement traces and eye traces were recorded, resulting in a total watching time of over 100 hours.

To align with the problem setting, we configured the time interval to one second. 
In other words, to represent the user's viewing content during each second, we extracted one frame from the $360^\circ$ video frames per second.
After collecting the video set, we then calculated the viewport position for each second based on the corresponding user head and eye traces.
Specifically, we mapped the head data onto the video frame coordinates, followed by the selection of the viewport whose geometric center is closest to the point as the ground truth.
In order to tackle the problem formulation, we selected the historical traces of continuous $t$ seconds and the video frames of $t + T$ seconds as the input sequence. 
The ground truth for the upcoming $T$ seconds of viewport positions was utilized.
We implemented a sliding-window mechanism with one-second to generate additional input sequences.
This implies that the start time of two adjacent input sequences varied by one second.
Finally, we randomly split the generated sequences into a test set, validation set, and training set, with a ratio of $1:1:8$. 

\paragraph{Metrics.} To evaluate the performance of our proposed method MFTR, we use average prediction accuracy (AP) which gives clearly quantitative results to measure the accuracy of predicted results. It is calculated by:
\begin{equation}
\begin{aligned}
    AP  =  \frac{1}{T}\sum_{i=1}^{T}( \hat{p}_{i}==p_{i}),
\end{aligned} 
\end{equation}
where $\hat{p_{i}}$ denotes the ground truth viewport and $p_{i}$ denotes predicted viewport at timestamp $i$. 

Following the previous work \cite{chao2021transformer, xu2018predicting, li2022spherical} we also use the the viewport overlap ratio metric (AO) which is the average percentage of the part of predicted viewports that inside the actual viewport. It can be calculated by:
\begin{equation}
\begin{aligned}
    AO  =  \frac{1}{T}\sum_{i=1}^{T}\frac{ S(\hat{p}_{i}, W, H) \cap S(p_{i}, W, H)} {S(\hat{p}_{i}, W, H)},
\end{aligned} 
\end{equation}
where $S()$ is function to calculate the viewport region given viewport center coordinate $p_t$, height $H$ and width $W$.

Similar to \cite{li2022spherical}, we incorporate delay time, which refers to the time required for prediction in one batch using the trained model weights.
Specifically, we set the prediction length $T$ to $5$ seconds and batch size to $16$.
Delay time could testify simplicity and computation efficiency of our method.

\paragraph{Implementation Details.} Our model consists of LSTMs with three hidden layers, each with a hidden size of 256, and transformers with a dimension of 512. To reduce computational complexity, we down-sampled the resolution of each selected 360-degree video frame from $3840 \times 2048$ to $720 /times 360$. 
Furthermore, we set the length of the historical sequence to five seconds (t=5) and the maximum prediction length to five seconds (T=5). 
We divided each 360-degree video frame into 10x20 tiles, with each viewport containing $4 \times 9$ tiles. During training, we set $\alpha$ and $\beta$ to 0.35 and 0.65, respectively. We used the AdamW optimizer with a learning rate of $10^{-3}$ and implemented early stopping on the validation set if the accuracy did not improve over ten epochs. The batch size was 16, and the model was trained for 200 epochs.
Both of the experiments were conducted on GTX-3090 platform, with each dataset requiring approximately two hours of training time.

\subsection{Performance Evaluation}
In order to evaluate the effectiveness of our proposed method MFTR, we compared it against three baseline studies: Offline-DHP \cite{xu2018predicting}, VPT360 \cite{chao2021transformer}, and SPVP360 \cite{li2022spherical}.
We opted to compare our approach with Offline-DHP, which is the most widely recognized option, as well as two of the most recent open-source state-of-the-art methods under the same data and settings.
It should be noted that SPVP360, a recently published method, has attained the state-of-the-art results in viewport prediction.

\begin{figure}[t]
\setlength{\abovecaptionskip}{0.cm}
    \centering
            \subfigure[AP comparison on PVS-HM]{
        \centering
        \includegraphics[width=0.47\linewidth]{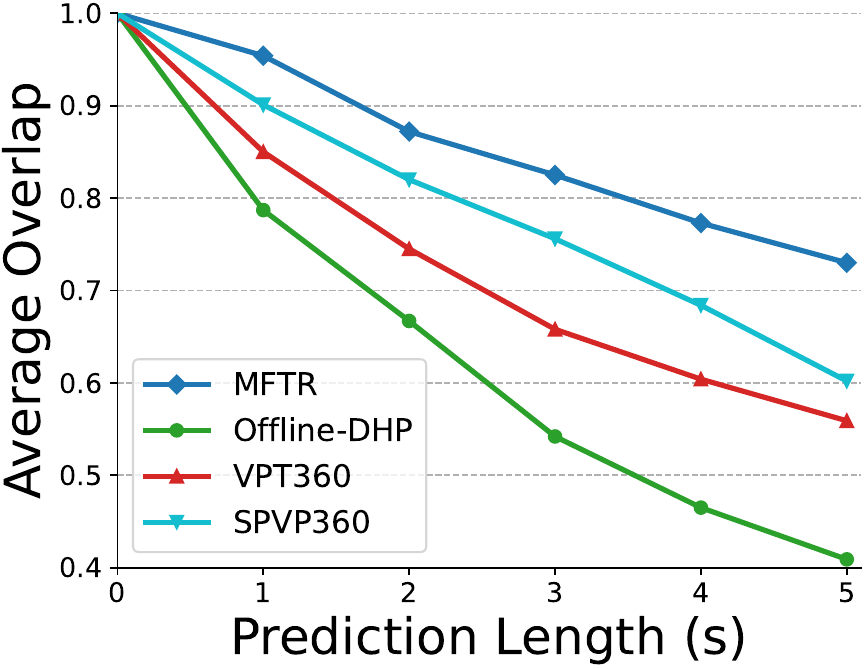}
        \label{dataset-result-1}
        } 
    \subfigure[AO comparison on PVS-HM]{
        \centering
        \includegraphics[width=0.47\linewidth]{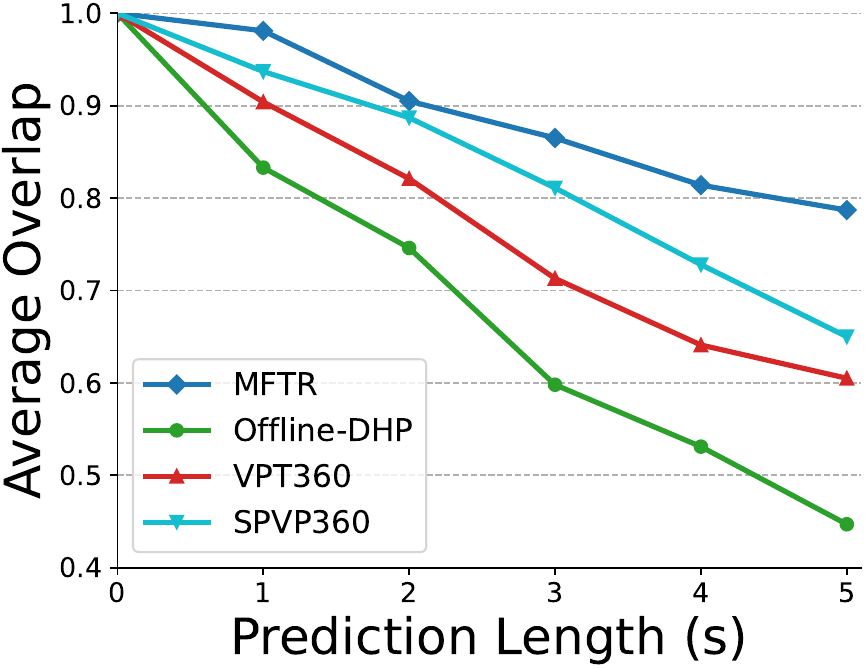}
        \label{dataset-result-2}
        } 
    \subfigure[AP comparison on Xu-Gaze]{
        \centering
        \includegraphics[width=0.47\linewidth]{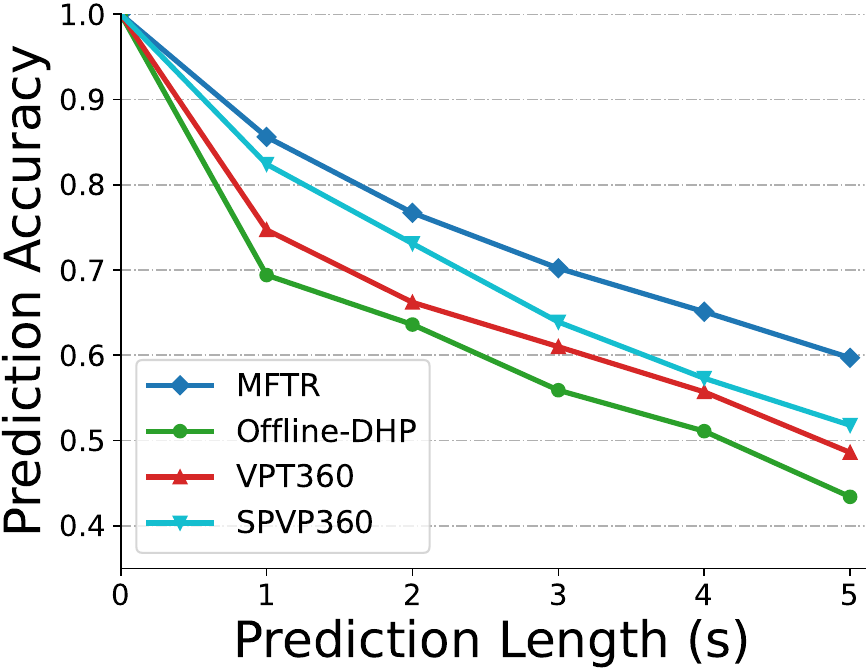}
        \label{dataset-result-3}
        } 
    \subfigure[AO comparison on Xu-Gaze]{
        \centering
        \includegraphics[width=0.47\linewidth]{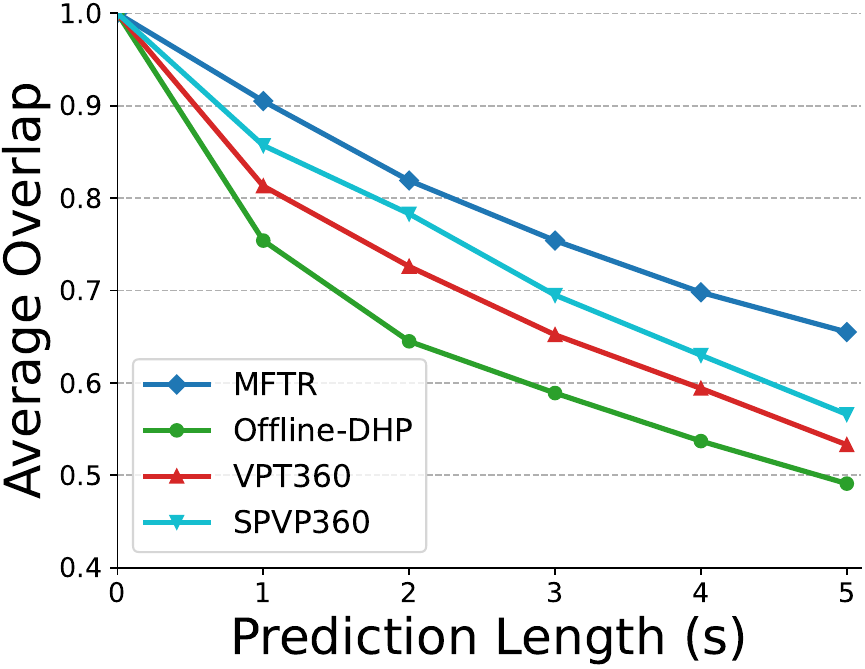}
        \label{dataset-result-4}
        } 

    \caption{Average precision (AP) and Average overlap (AO) comparison with baseline methods on PVS-HM and XuGaze dataset.}

\end{figure}

\paragraph{Results on PVS-HM dataset.} 
For the prediction of future viewports, we employed the past five seconds of user trajectories and video contents, with prediction lengths ranging from one second to five seconds.
Table \ref{fov_accuracy_table} compares the AP and AO metrics of our proposed MFTR method with selected state-of-the-art methods. 
And the result is represented in line format in Fig. \ref{dataset-result-1} and Fig. \ref{dataset-result-2}.
Our results indicate that:
i) MFTR outperforms the state-of-the-art methods by a significant margin. Specifically, the overall average prediction accuracy (AP) and average overlap ratio (AO) improved by $7.82\%$ and $6.78\%$, respectively, even when compared to the best baseline method SPVP360. This improvement can be attributed to the proposed tile-guided prediction mechanism, which reduces fluctuations and provides a more robust viewport selection mechanism.
ii) As a metric for model stability, we propose utilizing the percentage reduction in prediction time from one second to five seconds.
Compared to the previous state-of-the-art method SPVP360, MFTR shows stability improvements of $7.50\%$ and $9.30\%$ in terms of AP and AO respectively. 
This result indicates our transformer-based framework can effectively explore long-range dependencies in single modality and successfully model interactions between different modalities.

\paragraph{Results on Xu-Gaze dataset.} We conducted experiments on the Xu-Gaze dataset using the same training setting as on the PVS-HM dataset to further validate our results. As depicted in Table \ref{fov_accuracy_table_xu}, Fig. \ref{dataset-result-3} and Fig. \ref{dataset-result-4}, our proposed method MFTR still achieves significant improvements in terms of average prediction accuracy (AP) and overlap ratio, with gains of $5.76\%$ and $6.00\%$, respectively, compared to the baseline method SPVP360. Furthermore, our method MFTR exhibits improved stability over long-term predictions with gains of $4.60\%$ and $1.99\%$ in terms of AP and AO, respectively. The transferability of our model to a different dataset such as Xu-Gaze confirms the effectiveness of our approach.

\begin{table}[t]

    \centering
    \resizebox{.91\linewidth}{!}{
    \begin{tabular}{c|c|c|c|c}
    \toprule  
    Method & Offline-DHP & VPT360& SPVP360 & MFTR\\
    \midrule  
    Delay(ms)& 102& \textbf{21}& 124 &74\\
    \midrule  
    AP(\%) &40.9& 55.9& 60.2& \textbf{73.0}\\
    \bottomrule 
    \end{tabular}}
    \caption{Comparison of delay time and average precision (AP) with baseline methods.}
    \label{cost}  
    \vspace{-0.5cm}
\end{table}

\paragraph{Computation efficiency.} 
In order to demonstrate the computational effectiveness of our approach MFTR, we conducted delay time experiments on GTX-3090 platform given trained model.
We measured the prediction time in one batch to represent delay, where we set the prediction length to 5 seconds and batch size to 16.
We compared our delay time and average prediction accuracy (AP) with baseline methods on the PVS-HM dataset. 
Table \ref{cost} shows that our MFTR method achieves the best AP performance, with a delay time that is only slightly higher than that of VPT360 (which considers only user's historical temporal traces) and much lower than that of Offline-DHP and SPVP360. 
Our experiment on delay time serves as evidence of the computational efficiency and simplicity of our model, reducing the likelihood of model fitting and improving the efficiency of video transmission.

\begin{figure*}[t]
  \centering
  \includegraphics[width=0.9\linewidth]{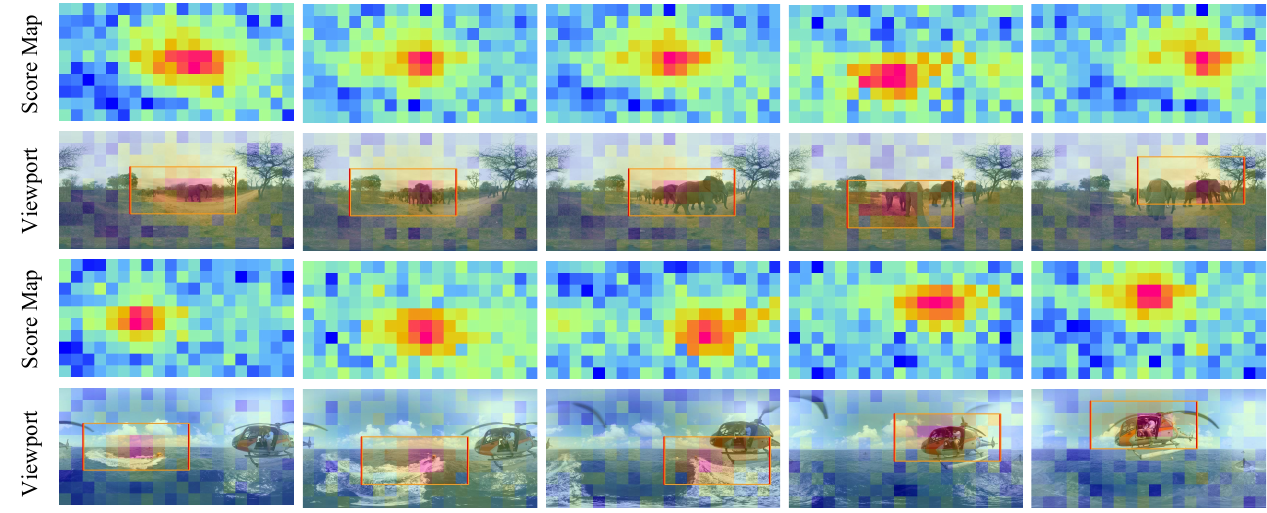}
  \caption{Visualization of score map and corresponding predicted viewport. 1st and 3rd rows depict the generated scores of each tile to represent how interested user is in. 2nd and 4th rows are corresponding determined viewport. We can observe that the selected viewports (red bounding boxes) and ground-truth viewports (yellow bounding boxes) are perfectly coincided.}
  \label{visualization}
  \vspace{-0.3cm}
\end{figure*}

\subsection{Ablation Study}
To verify the effectiveness of designs in our proposed framework, we conduct a series of ablation experiments on PVS-HM dataset and report the results of average prediction accuracy (AP) as the main indicator. 

\begin{table}[t]
    \centering
    \resizebox{1\linewidth}{!}{
    \begin{tabular}{c|ccccc}
    \toprule  
    Prediction length& 1s & 2s &3s & 4s &5s\\
    \midrule  
    w/o Temporal Transformer& 0.741& 0.689& 0.573& 0.504& 0.436\\
    w/o Position Prediction Head& 0.890& 0.822& 0.771& 0.673& 0.639\\
    w/o Visual Transformer& 0.869& 0.803& 0.751& 0.671& 0.627\\
    w/o Temporal-Visual Fusion & 0.904& 0.847& 0.779& 0.684&0.633\\
    w/o Tile Classification Head & 0.881& 0.795& 0.755& 0.702& 0.642\\
    
    \midrule  
     MFTR&\textbf{0.954}& \textbf{0.872}& \textbf{0.825}& \textbf{0.773}& \textbf{0.730}\\
    \bottomrule 
    \end{tabular}}
    
    \caption{Ablation studies of different modules.}
    \label{ablation}
\end{table}

\begin{table}[t]
\vspace{-0.5cm}
    \centering
    \resizebox{1\linewidth}{!}{
    \begin{tabular}{c|ccccc}
    \toprule  
    Prediction length& 1s & 2s &3s & 4s &5s\\
    \midrule  
    2 layers& 0.811& 0.766& 0.681& 0.601& 0.514\\
    4 layers& 0.933& 0.851& 0.796& 0.720& 0.674\\
    6 layers&\textbf{0.954}& \textbf{0.872}& \textbf{0.825}& \textbf{0.773}& \textbf{0.730} \\
    8 layers& 0.894& 0.825& 0.731& 0.671& 0.580\\
    \bottomrule 
    \end{tabular}}
    
    \caption{Prediction performance on PVS-HM dataset of our method with different number of encoder layers.}
    \label{attention-layers}
    \vspace{-0.5cm}
\end{table}

\begin{table}[t]
    \centering
    \resizebox{1\linewidth}{!}{
    \begin{tabular}{c|c|c|c|ccccc}
    \toprule  
    \multicolumn{4}{c|}{Prediction length}& 1s & 2s &3s & 4s &5s\\
    \midrule  
    \multirow{5}{*}{$\alpha$}& 0.25& \multirow{5}{*}{$\beta$}& 0.75& 0.892& 0.849& 0.782& 0.703& 0.645\\
    & 0.30& &  0.70& 0.912& 0.854& 0.811& 0.748& 0.673\\
    & \textbf{0.35}& & \textbf{0.65}& \textbf{0.954}& \textbf{0.872}& \textbf{0.825}& \textbf{0.773}& \textbf{0.730} \\
    &0.40 & &0.60& 0.921& 0.850& 0.809& 0.756& 0.704\\
    &0.45 & &0.55& 0.874& 0.805& 0.752& 0.698& 0.633\\
    \bottomrule 
    \end{tabular}}
    \caption{Prediction performance of our method on PVS-HM dataset with different hyper-parameters $\alpha$, $\beta$.}
    \label{hyper} 
    \vspace{-0.8cm}
\end{table}

\paragraph{Effects of different components.} As shown in Table \ref{ablation}, we present experimental results to measure the contributions of each component in MFTR. 
Our results indicate that the absence of the Temporal Transformer leads to a significant decrease in model performance, underscoring the efficacy and robustness of our Temporal Transformer in capturing user motion patterns.
Additionally, the removal of the Position Prediction Head results in a performance drop of $7.18\%$, which underscores the significance of this component in training temporal feature construction, as well as the efficacy of utilizing $\mathcal{L}_{cls}$ to supervise model training.

The absence of the Visual Transformer results in an $8.66\%$ decrease in the overall prediction accuracy, indicating the transformer-generated sequential visual features from $360^{\circ}$ video frames indeed explore the dependencies within the visual modality and improve the performance. Moreover, the Temporal-Visual Fusion Module improves prediction stability by $6.14\%$, emphasizing its importance in modeling the impact of both temporal and visual modalities and mining inter-modality relations. Lastly, we evaluate the effectiveness of our Tile Selection Head by comparing it with the method of constructing viewports through predicted head movements. The Tile Selection Head significantly enhances viewport accuracy by $7.58\%$, demonstrating the efficiency of our tile classification-based prediction mechanism compared to viewport prediction tasks.

\paragraph{Effects of encoder layer numbers.} 
We conducted experiments to evaluate the impact of the multi-head attention mechanism by varying the number of encoder layers in the transformer structures. The results are presented in Table \ref{attention-layers}, where we observe that increasing the number of encoder layers from $2$ to $6$ improves the average prediction accuracy (AP) by $15.62\%$, while further increasing the number to $8$ causes a drop of $9.06\%$ compared to $6$ layers. This suggests that the performance of the model is not always positively correlated with the number of encoder layers, and that a transformer with $6$ encoder layers can effectively enhance single-modal features and capture inter- and intra-modality relations.

\paragraph{Effects of $\alpha$ and $\beta$.} The hyper-parameters $\alpha$, $\beta$ in Equation \ref{loss} are used to balance fusion process within the temporal modality, as well as between different modalities.
If the proportion of head prediction in the gradient return is reduced by setting $\alpha$ to a small value, it leads to incomplete temporal representations and ultimately reduces the overall performance of our method.
Likewise, if $\beta$ is set to a small value, the tile classification process may be incomplete and the multi-modal fusion may not be fully achieved in depth.
To determine the optimal hyper-parameter, we train our model with various combinations as shown in Table \ref{hyper}, $\alpha$ = 0.35 and $\beta$ = 0.65 provides a great balance and achieves best performance for MFTR.
It appears that the weight of $\beta$ is higher, because $\mathcal{L}_{cls}$ promotes global convergence, serves as a general measure for evaluating the impact of temporal-visual features.
While $\mathcal{L}_{pos}$ is more locally, specifically utilized to facilitate fusion process of temporal features.

\subsection{Visualization}
Fig. \ref{visualization} provides qualitative examples of tile classification results and the corresponding determined viewports, allowing for a deeper insight into our method. Heatmaps of classification scores $S^\in\mathbb{R}^{N_h\times N_w}$ for all tiles in each selected $360^{\circ}$ video frame are drawn, and chosen viewport regions based on tile classification results are presented. Our results demonstrate that our MFTR model accurately distinguishes the tiles of user interest and predicts viewports (red bounding boxes) that perfectly coincide with ground-truth viewports (yellow bounding boxes), explicitly demonstrating the effectiveness of our method. Additionally, our viewport selection method based on the score map from tile classification is more interpretable when compared with trajectory-based approaches.

\section{Conclusion}
In this work, we have developed a tile classification based viewport prediction method with Multi-modal Fusion Transformer (MFTR), which novelly transfers viewport prediction to a tile classification task. 
MFTR first extracts long-range dependencies within the same modality through transformers. 
Then it fuses different modalities into cross-modal representation to model the combined influence of user historical inputs and video contents. 
Moreover, MFTR classifies the future tiles into two categories: user interested tile and not. 
The future viewport could be determined in a more robust manner of selecting regions containing most tiles of user interest. 
The experimental results indicates that our proposed method MFTR significantly surpasses the state-of-the-art methods with higher prediction accuracy and competitive computation efficiency.

\bibliographystyle{ACM-Reference-Format}
\balance
\bibliography{viewport}


\begin{thebibliography}{43}


\ifx \showCODEN    \undefined \def \showCODEN     #1{\unskip}     \fi
\ifx \showDOI      \undefined \def \showDOI       #1{#1}\fi
\ifx \showISBNx    \undefined \def \showISBNx     #1{\unskip}     \fi
\ifx \showISBNxiii \undefined \def \showISBNxiii  #1{\unskip}     \fi
\ifx \showISSN     \undefined \def \showISSN      #1{\unskip}     \fi
\ifx \showLCCN     \undefined \def \showLCCN      #1{\unskip}     \fi
\ifx \shownote     \undefined \def \shownote      #1{#1}          \fi
\ifx \showarticletitle \undefined \def \showarticletitle #1{#1}   \fi
\ifx \showURL      \undefined \def \showURL       {\relax}        \fi
\providecommand\bibfield[2]{#2}
\providecommand\bibinfo[2]{#2}
\providecommand\natexlab[1]{#1}
\providecommand\showeprint[2][]{arXiv:#2}

\bibitem[Aladagli et~al\mbox{.}(2017)]%
        {aladagli2017predicting}
\bibfield{author}{\bibinfo{person}{A~Deniz Aladagli}, \bibinfo{person}{Erhan Ekmekcioglu}, \bibinfo{person}{Dmitri Jarnikov}, {and} \bibinfo{person}{Ahmet Kondoz}.} \bibinfo{year}{2017}\natexlab{}.
\newblock \showarticletitle{Predicting head trajectories in 360 virtual reality videos}. In \bibinfo{booktitle}{\emph{2017 International Conference on 3D Immersion (IC3D)}}. IEEE, \bibinfo{pages}{1--6}.
\newblock


\bibitem[Ban et~al\mbox{.}(2018)]%
        {ban2018cub360}
\bibfield{author}{\bibinfo{person}{Yixuan Ban}, \bibinfo{person}{Lan Xie}, \bibinfo{person}{Zhimin Xu}, \bibinfo{person}{Xinggong Zhang}, \bibinfo{person}{Zongming Guo}, {and} \bibinfo{person}{Yue Wang}.} \bibinfo{year}{2018}\natexlab{}.
\newblock \showarticletitle{Cub360: Exploiting cross-users behaviors for viewport prediction in 360 video adaptive streaming}. In \bibinfo{booktitle}{\emph{2018 IEEE International Conference on Multimedia and Expo (ICME)}}. IEEE, \bibinfo{pages}{1--6}.
\newblock


\bibitem[Carion et~al\mbox{.}(2020)]%
        {carion2020end}
\bibfield{author}{\bibinfo{person}{Nicolas Carion}, \bibinfo{person}{Francisco Massa}, \bibinfo{person}{Gabriel Synnaeve}, \bibinfo{person}{Nicolas Usunier}, \bibinfo{person}{Alexander Kirillov}, {and} \bibinfo{person}{Sergey Zagoruyko}.} \bibinfo{year}{2020}\natexlab{}.
\newblock \showarticletitle{End-to-end object detection with transformers}. In \bibinfo{booktitle}{\emph{European conference on computer vision}}. Springer, \bibinfo{pages}{213--229}.
\newblock


\bibitem[Chao et~al\mbox{.}(2021)]%
        {chao2021transformer}
\bibfield{author}{\bibinfo{person}{Fang-Yi Chao}, \bibinfo{person}{Cagri Ozcinar}, {and} \bibinfo{person}{Aljosa Smolic}.} \bibinfo{year}{2021}\natexlab{}.
\newblock \showarticletitle{Transformer-based Long-Term Viewport Prediction in 360° Video: Scanpath is All You Need.}. In \bibinfo{booktitle}{\emph{MMSP}}. \bibinfo{pages}{1--6}.
\newblock


\bibitem[Chen et~al\mbox{.}(2020)]%
        {chen2020sparkle}
\bibfield{author}{\bibinfo{person}{Jinyu Chen}, \bibinfo{person}{Xianzhuo Luo}, \bibinfo{person}{Miao Hu}, \bibinfo{person}{Di Wu}, {and} \bibinfo{person}{Yipeng Zhou}.} \bibinfo{year}{2020}\natexlab{}.
\newblock \showarticletitle{Sparkle: User-aware viewport prediction in 360-degree video streaming}.
\newblock \bibinfo{journal}{\emph{IEEE Transactions on Multimedia}}  \bibinfo{volume}{23} (\bibinfo{year}{2020}), \bibinfo{pages}{3853--3866}.
\newblock


\bibitem[Chiariotti(2021)]%
        {chiariotti2021survey}
\bibfield{author}{\bibinfo{person}{Federico Chiariotti}.} \bibinfo{year}{2021}\natexlab{}.
\newblock \showarticletitle{A survey on 360-degree video: Coding, quality of experience and streaming}.
\newblock \bibinfo{journal}{\emph{Computer Communications}}  \bibinfo{volume}{177} (\bibinfo{year}{2021}), \bibinfo{pages}{133--155}.
\newblock


\bibitem[Chopra et~al\mbox{.}(2021)]%
        {chopra2021parima}
\bibfield{author}{\bibinfo{person}{Lovish Chopra}, \bibinfo{person}{Sarthak Chakraborty}, \bibinfo{person}{Abhijit Mondal}, {and} \bibinfo{person}{Sandip Chakraborty}.} \bibinfo{year}{2021}\natexlab{}.
\newblock \showarticletitle{Parima: Viewport adaptive 360-degree video streaming}. In \bibinfo{booktitle}{\emph{Proceedings of the Web Conference 2021}}. \bibinfo{pages}{2379--2391}.
\newblock


\bibitem[de~la Fuente et~al\mbox{.}(2019)]%
        {de2019delay}
\bibfield{author}{\bibinfo{person}{Yago~Sanchez de~la Fuente}, \bibinfo{person}{Gurdeep~Singh Bhullar}, \bibinfo{person}{Robert Skupin}, \bibinfo{person}{Cornelius Hellge}, {and} \bibinfo{person}{Thomas Schierl}.} \bibinfo{year}{2019}\natexlab{}.
\newblock \showarticletitle{Delay impact on MPEG OMAF’s tile-based viewport-dependent 360 video streaming}.
\newblock \bibinfo{journal}{\emph{IEEE Journal on Emerging and Selected Topics in Circuits and Systems}} \bibinfo{volume}{9}, \bibinfo{number}{1} (\bibinfo{year}{2019}), \bibinfo{pages}{18--28}.
\newblock


\bibitem[Devlin et~al\mbox{.}(2018)]%
        {devlin2018bert}
\bibfield{author}{\bibinfo{person}{Jacob Devlin}, \bibinfo{person}{Ming-Wei Chang}, \bibinfo{person}{Kenton Lee}, {and} \bibinfo{person}{Kristina Toutanova}.} \bibinfo{year}{2018}\natexlab{}.
\newblock \showarticletitle{Bert: Pre-training of deep bidirectional transformers for language understanding}.
\newblock \bibinfo{journal}{\emph{arXiv preprint arXiv:1810.04805}} (\bibinfo{year}{2018}).
\newblock


\bibitem[Dosovitskiy et~al\mbox{.}(2020)]%
        {dosovitskiy2020image}
\bibfield{author}{\bibinfo{person}{Alexey Dosovitskiy}, \bibinfo{person}{Lucas Beyer}, \bibinfo{person}{Alexander Kolesnikov}, \bibinfo{person}{Dirk Weissenborn}, \bibinfo{person}{Xiaohua Zhai}, \bibinfo{person}{Thomas Unterthiner}, \bibinfo{person}{Mostafa Dehghani}, \bibinfo{person}{Matthias Minderer}, \bibinfo{person}{Georg Heigold}, \bibinfo{person}{Sylvain Gelly}, {et~al\mbox{.}}} \bibinfo{year}{2020}\natexlab{}.
\newblock \showarticletitle{An image is worth 16x16 words: Transformers for image recognition at scale}.
\newblock \bibinfo{journal}{\emph{arXiv preprint arXiv:2010.11929}} (\bibinfo{year}{2020}).
\newblock


\bibitem[Feng et~al\mbox{.}(2019)]%
        {feng2019viewport}
\bibfield{author}{\bibinfo{person}{Xianglong Feng}, \bibinfo{person}{Viswanathan Swaminathan}, {and} \bibinfo{person}{Sheng Wei}.} \bibinfo{year}{2019}\natexlab{}.
\newblock \showarticletitle{Viewport prediction for live 360-degree mobile video streaming using user-content hybrid motion tracking}.
\newblock \bibinfo{journal}{\emph{Proceedings of the ACM on Interactive, Mobile, Wearable and Ubiquitous Technologies}} \bibinfo{volume}{3}, \bibinfo{number}{2} (\bibinfo{year}{2019}), \bibinfo{pages}{1--22}.
\newblock


\bibitem[Han et~al\mbox{.}(2022)]%
        {han2022survey}
\bibfield{author}{\bibinfo{person}{Kai Han}, \bibinfo{person}{Yunhe Wang}, \bibinfo{person}{Hanting Chen}, \bibinfo{person}{Xinghao Chen}, \bibinfo{person}{Jianyuan Guo}, \bibinfo{person}{Zhenhua Liu}, \bibinfo{person}{Yehui Tang}, \bibinfo{person}{An Xiao}, \bibinfo{person}{Chunjing Xu}, \bibinfo{person}{Yixing Xu}, {et~al\mbox{.}}} \bibinfo{year}{2022}\natexlab{}.
\newblock \showarticletitle{A survey on vision transformer}.
\newblock \bibinfo{journal}{\emph{IEEE transactions on pattern analysis and machine intelligence}} \bibinfo{volume}{45}, \bibinfo{number}{1} (\bibinfo{year}{2022}), \bibinfo{pages}{87--110}.
\newblock


\bibitem[Jamali et~al\mbox{.}(2020)]%
        {jamali2020lstm}
\bibfield{author}{\bibinfo{person}{Mohammadreza Jamali}, \bibinfo{person}{St{\'e}phane Coulombe}, \bibinfo{person}{Ahmad Vakili}, {and} \bibinfo{person}{Carlos Vazquez}.} \bibinfo{year}{2020}\natexlab{}.
\newblock \showarticletitle{LSTM-based viewpoint prediction for multi-quality tiled video coding in virtual reality streaming}. In \bibinfo{booktitle}{\emph{2020 IEEE International Symposium on Circuits and Systems (ISCAS)}}. IEEE, \bibinfo{pages}{1--5}.
\newblock


\bibitem[Jiang et~al\mbox{.}(2020)]%
        {jiang2020svp}
\bibfield{author}{\bibinfo{person}{Xiaolan Jiang}, \bibinfo{person}{Si~Ahmed Naas}, \bibinfo{person}{Yi-Han Chiang}, \bibinfo{person}{Stephan Sigg}, {and} \bibinfo{person}{Yusheng Ji}.} \bibinfo{year}{2020}\natexlab{}.
\newblock \showarticletitle{SVP: Sinusoidal viewport prediction for 360-degree video streaming}.
\newblock \bibinfo{journal}{\emph{IEEE Access}}  \bibinfo{volume}{8} (\bibinfo{year}{2020}), \bibinfo{pages}{164471--164481}.
\newblock


\bibitem[Jiang et~al\mbox{.}(2022)]%
        {jiang2022robust}
\bibfield{author}{\bibinfo{person}{Yuang Jiang}, \bibinfo{person}{Konstantinos Poularakis}, \bibinfo{person}{Diego Kiedanski}, \bibinfo{person}{Sastry Kompella}, {and} \bibinfo{person}{Leandros Tassiulas}.} \bibinfo{year}{2022}\natexlab{}.
\newblock \showarticletitle{Robust and Resource-efficient Machine Learning Aided Viewport Prediction in Virtual Reality}.
\newblock \bibinfo{journal}{\emph{arXiv preprint arXiv:2212.09945}} (\bibinfo{year}{2022}).
\newblock


\bibitem[Kang et~al\mbox{.}(2020)]%
        {kang2020natural}
\bibfield{author}{\bibinfo{person}{Yue Kang}, \bibinfo{person}{Zhao Cai}, \bibinfo{person}{Chee-Wee Tan}, \bibinfo{person}{Qian Huang}, {and} \bibinfo{person}{Hefu Liu}.} \bibinfo{year}{2020}\natexlab{}.
\newblock \showarticletitle{Natural language processing (NLP) in management research: A literature review}.
\newblock \bibinfo{journal}{\emph{Journal of Management Analytics}} \bibinfo{volume}{7}, \bibinfo{number}{2} (\bibinfo{year}{2020}), \bibinfo{pages}{139--172}.
\newblock


\bibitem[Kim et~al\mbox{.}(2021)]%
        {kim2021vilt}
\bibfield{author}{\bibinfo{person}{Wonjae Kim}, \bibinfo{person}{Bokyung Son}, {and} \bibinfo{person}{Ildoo Kim}.} \bibinfo{year}{2021}\natexlab{}.
\newblock \showarticletitle{Vilt: Vision-and-language transformer without convolution or region supervision}. In \bibinfo{booktitle}{\emph{International Conference on Machine Learning}}. PMLR, \bibinfo{pages}{5583--5594}.
\newblock


\bibitem[Lee et~al\mbox{.}(2021)]%
        {lee2021prediction}
\bibfield{author}{\bibinfo{person}{Dongwon Lee}, \bibinfo{person}{Minji Choi}, {and} \bibinfo{person}{Joohyun Lee}.} \bibinfo{year}{2021}\natexlab{}.
\newblock \showarticletitle{Prediction of head movement in 360-degree videos using attention model}.
\newblock \bibinfo{journal}{\emph{Sensors}} \bibinfo{volume}{21}, \bibinfo{number}{11} (\bibinfo{year}{2021}), \bibinfo{pages}{3678}.
\newblock


\bibitem[Li et~al\mbox{.}(2019)]%
        {li2019very}
\bibfield{author}{\bibinfo{person}{Chenge Li}, \bibinfo{person}{Weixi Zhang}, \bibinfo{person}{Yong Liu}, {and} \bibinfo{person}{Yao Wang}.} \bibinfo{year}{2019}\natexlab{}.
\newblock \showarticletitle{Very long term field of view prediction for 360-degree video streaming}. In \bibinfo{booktitle}{\emph{2019 IEEE Conference on Multimedia Information Processing and Retrieval (MIPR)}}. IEEE, \bibinfo{pages}{297--302}.
\newblock


\bibitem[Li et~al\mbox{.}(2020)]%
        {li2020unicoder}
\bibfield{author}{\bibinfo{person}{Gen Li}, \bibinfo{person}{Nan Duan}, \bibinfo{person}{Yuejian Fang}, \bibinfo{person}{Ming Gong}, {and} \bibinfo{person}{Daxin Jiang}.} \bibinfo{year}{2020}\natexlab{}.
\newblock \showarticletitle{Unicoder-vl: A universal encoder for vision and language by cross-modal pre-training}. In \bibinfo{booktitle}{\emph{Proceedings of the AAAI Conference on Artificial Intelligence}}, Vol.~\bibinfo{volume}{34}. \bibinfo{pages}{11336--11344}.
\newblock


\bibitem[Li et~al\mbox{.}(2022)]%
        {li2022spherical}
\bibfield{author}{\bibinfo{person}{Jie Li}, \bibinfo{person}{Ling Han}, \bibinfo{person}{Chong Zhang}, \bibinfo{person}{Qiyue Li}, {and} \bibinfo{person}{Zhi Liu}.} \bibinfo{year}{2022}\natexlab{}.
\newblock \showarticletitle{Spherical Convolution empowered Viewport Prediction in 360 Video Multicast with Limited FoV Feedback}.
\newblock \bibinfo{journal}{\emph{ACM Transactions on Multimedia Computing, Communications, and Applications (TOMM)}} (\bibinfo{year}{2022}).
\newblock


\bibitem[Liu et~al\mbox{.}(2021)]%
        {liu2021swin}
\bibfield{author}{\bibinfo{person}{Ze Liu}, \bibinfo{person}{Yutong Lin}, \bibinfo{person}{Yue Cao}, \bibinfo{person}{Han Hu}, \bibinfo{person}{Yixuan Wei}, \bibinfo{person}{Zheng Zhang}, \bibinfo{person}{Stephen Lin}, {and} \bibinfo{person}{Baining Guo}.} \bibinfo{year}{2021}\natexlab{}.
\newblock \showarticletitle{Swin transformer: Hierarchical vision transformer using shifted windows}. In \bibinfo{booktitle}{\emph{Proceedings of the IEEE/CVF international conference on computer vision}}. \bibinfo{pages}{10012--10022}.
\newblock


\bibitem[Liu et~al\mbox{.}(2022)]%
        {liu2022video}
\bibfield{author}{\bibinfo{person}{Ze Liu}, \bibinfo{person}{Jia Ning}, \bibinfo{person}{Yue Cao}, \bibinfo{person}{Yixuan Wei}, \bibinfo{person}{Zheng Zhang}, \bibinfo{person}{Stephen Lin}, {and} \bibinfo{person}{Han Hu}.} \bibinfo{year}{2022}\natexlab{}.
\newblock \showarticletitle{Video swin transformer}. In \bibinfo{booktitle}{\emph{Proceedings of the IEEE/CVF conference on computer vision and pattern recognition}}. \bibinfo{pages}{3202--3211}.
\newblock


\bibitem[Lu et~al\mbox{.}(2019)]%
        {lu2019vilbert}
\bibfield{author}{\bibinfo{person}{Jiasen Lu}, \bibinfo{person}{Dhruv Batra}, \bibinfo{person}{Devi Parikh}, {and} \bibinfo{person}{Stefan Lee}.} \bibinfo{year}{2019}\natexlab{}.
\newblock \showarticletitle{Vilbert: Pretraining task-agnostic visiolinguistic representations for vision-and-language tasks}.
\newblock \bibinfo{journal}{\emph{Advances in neural information processing systems}}  \bibinfo{volume}{32} (\bibinfo{year}{2019}).
\newblock


\bibitem[Lu et~al\mbox{.}(2020)]%
        {lu202012}
\bibfield{author}{\bibinfo{person}{Jiasen Lu}, \bibinfo{person}{Vedanuj Goswami}, \bibinfo{person}{Marcus Rohrbach}, \bibinfo{person}{Devi Parikh}, {and} \bibinfo{person}{Stefan Lee}.} \bibinfo{year}{2020}\natexlab{}.
\newblock \showarticletitle{12-in-1: Multi-task vision and language representation learning}. In \bibinfo{booktitle}{\emph{Proceedings of the IEEE/CVF Conference on Computer Vision and Pattern Recognition}}. \bibinfo{pages}{10437--10446}.
\newblock


\bibitem[Nasrabadi et~al\mbox{.}(2020)]%
        {nasrabadi2020viewport}
\bibfield{author}{\bibinfo{person}{Afshin~Taghavi Nasrabadi}, \bibinfo{person}{Aliehsan Samiei}, {and} \bibinfo{person}{Ravi Prakash}.} \bibinfo{year}{2020}\natexlab{}.
\newblock \showarticletitle{Viewport prediction for 360 videos: a clustering approach}. In \bibinfo{booktitle}{\emph{Proceedings of the 30th ACM Workshop on Network and Operating Systems Support for Digital Audio and Video}}. \bibinfo{pages}{34--39}.
\newblock


\bibitem[Nguyen et~al\mbox{.}(2018)]%
        {nguyen2018your}
\bibfield{author}{\bibinfo{person}{Anh Nguyen}, \bibinfo{person}{Zhisheng Yan}, {and} \bibinfo{person}{Klara Nahrstedt}.} \bibinfo{year}{2018}\natexlab{}.
\newblock \showarticletitle{Your attention is unique: Detecting 360-degree video saliency in head-mounted display for head movement prediction}. In \bibinfo{booktitle}{\emph{Proceedings of the 26th ACM international conference on Multimedia}}. \bibinfo{pages}{1190--1198}.
\newblock


\bibitem[Raffel et~al\mbox{.}(2019)]%
        {raffel2019t2t}
\bibfield{author}{\bibinfo{person}{Colin Raffel}, \bibinfo{person}{Noam Shazeer}, \bibinfo{person}{Adam Roberts}, \bibinfo{person}{Katherine Lee}, \bibinfo{person}{Sharan Narang}, \bibinfo{person}{Michael Matena}, \bibinfo{person}{Yanqi Zhou}, \bibinfo{person}{Wei Li}, \bibinfo{person}{Peter~J Liu}, {et~al\mbox{.}}} \bibinfo{year}{2019}\natexlab{}.
\newblock \showarticletitle{Exploring the limits of transfer learning with a unified text-to-text transformer}.
\newblock \bibinfo{journal}{\emph{arXiv preprint arXiv:1910.10683}} (\bibinfo{year}{2019}).
\newblock


\bibitem[Rondon et~al\mbox{.}(2019)]%
        {rondon2019revisiting}
\bibfield{author}{\bibinfo{person}{Miguel Fabian~Romero Rondon}, \bibinfo{person}{Lucile Sassatelli}, \bibinfo{person}{Ramon~Aparicio Pardo}, {and} \bibinfo{person}{Frederic Precioso}.} \bibinfo{year}{2019}\natexlab{}.
\newblock \showarticletitle{Revisiting Deep Architectures for Head Motion Prediction in 360 $\{$$\backslash$deg$\}$ Videos}.
\newblock \bibinfo{journal}{\emph{arXiv preprint arXiv:1911.11702}} (\bibinfo{year}{2019}).
\newblock


\bibitem[Sandler et~al\mbox{.}(2018)]%
        {sandler2018mobilenetv2}
\bibfield{author}{\bibinfo{person}{Mark Sandler}, \bibinfo{person}{Andrew Howard}, \bibinfo{person}{Menglong Zhu}, \bibinfo{person}{Andrey Zhmoginov}, {and} \bibinfo{person}{Liang-Chieh Chen}.} \bibinfo{year}{2018}\natexlab{}.
\newblock \showarticletitle{Mobilenetv2: Inverted residuals and linear bottlenecks}. In \bibinfo{booktitle}{\emph{Proceedings of the IEEE conference on computer vision and pattern recognition}}. \bibinfo{pages}{4510--4520}.
\newblock


\bibitem[Van~Damme et~al\mbox{.}(2022)]%
        {van2022machine}
\bibfield{author}{\bibinfo{person}{Sam Van~Damme}, \bibinfo{person}{Maria~Torres Vega}, {and} \bibinfo{person}{Filip De~Turck}.} \bibinfo{year}{2022}\natexlab{}.
\newblock \showarticletitle{Machine Learning based Content-Agnostic Viewport Prediction for 360-Degree Video}.
\newblock \bibinfo{journal}{\emph{ACM Transactions on Multimedia Computing, Communications, and Applications (TOMM)}} \bibinfo{volume}{18}, \bibinfo{number}{2} (\bibinfo{year}{2022}), \bibinfo{pages}{1--24}.
\newblock


\bibitem[van~der Hooft et~al\mbox{.}(2019)]%
        {van2019optimizing}
\bibfield{author}{\bibinfo{person}{Jeroen van~der Hooft}, \bibinfo{person}{Maria~Torres Vega}, \bibinfo{person}{Stefano Petrangeli}, \bibinfo{person}{Tim Wauters}, {and} \bibinfo{person}{Filip De~Turck}.} \bibinfo{year}{2019}\natexlab{}.
\newblock \showarticletitle{Optimizing adaptive tile-based virtual reality video streaming}. In \bibinfo{booktitle}{\emph{2019 IFIP/IEEE Symposium on Integrated Network and Service Management (IM)}}. IEEE, \bibinfo{pages}{381--387}.
\newblock


\bibitem[Vaswani et~al\mbox{.}(2017)]%
        {vaswani2017transformer}
\bibfield{author}{\bibinfo{person}{Ashish Vaswani}, \bibinfo{person}{Noam Shazeer}, \bibinfo{person}{Niki Parmar}, \bibinfo{person}{Jakob Uszkoreit}, \bibinfo{person}{Llion Jones}, \bibinfo{person}{Aidan~N Gomez}, \bibinfo{person}{\L~ukasz Kaiser}, {and} \bibinfo{person}{Illia Polosukhin}.} \bibinfo{year}{2017}\natexlab{}.
\newblock \showarticletitle{Attention is all you need}. In \bibinfo{booktitle}{\emph{Advances in Neural Information Processing Systems}}, Vol.~\bibinfo{volume}{30}.
\newblock


\bibitem[Xie et~al\mbox{.}(2017)]%
        {xie2017360probdash}
\bibfield{author}{\bibinfo{person}{Lan Xie}, \bibinfo{person}{Zhimin Xu}, \bibinfo{person}{Yixuan Ban}, \bibinfo{person}{Xinggong Zhang}, {and} \bibinfo{person}{Zongming Guo}.} \bibinfo{year}{2017}\natexlab{}.
\newblock \showarticletitle{360probdash: Improving qoe of 360 video streaming using tile-based http adaptive streaming}. In \bibinfo{booktitle}{\emph{Proceedings of the 25th ACM international conference on Multimedia}}. \bibinfo{pages}{315--323}.
\newblock


\bibitem[Xu et~al\mbox{.}(2018b)]%
        {xu2018predicting}
\bibfield{author}{\bibinfo{person}{Mai Xu}, \bibinfo{person}{Yuhang Song}, \bibinfo{person}{Jianyi Wang}, \bibinfo{person}{MingLang Qiao}, \bibinfo{person}{Liangyu Huo}, {and} \bibinfo{person}{Zulin Wang}.} \bibinfo{year}{2018}\natexlab{b}.
\newblock \showarticletitle{Predicting head movement in panoramic video: A deep reinforcement learning approach}.
\newblock \bibinfo{journal}{\emph{IEEE transactions on pattern analysis and machine intelligence}} \bibinfo{volume}{41}, \bibinfo{number}{11} (\bibinfo{year}{2018}), \bibinfo{pages}{2693--2708}.
\newblock


\bibitem[Xu et~al\mbox{.}(2019)]%
        {xu2019analyzing}
\bibfield{author}{\bibinfo{person}{Tan Xu}, \bibinfo{person}{Bo Han}, {and} \bibinfo{person}{Feng Qian}.} \bibinfo{year}{2019}\natexlab{}.
\newblock \showarticletitle{Analyzing viewport prediction under different VR interactions}. In \bibinfo{booktitle}{\emph{Proceedings of the 15th International Conference on Emerging Networking Experiments And Technologies}}. \bibinfo{pages}{165--171}.
\newblock


\bibitem[Xu et~al\mbox{.}(2018a)]%
        {xu2018gaze}
\bibfield{author}{\bibinfo{person}{Yanyu Xu}, \bibinfo{person}{Yanbing Dong}, \bibinfo{person}{Junru Wu}, \bibinfo{person}{Zhengzhong Sun}, \bibinfo{person}{Zhiru Shi}, \bibinfo{person}{Jingyi Yu}, {and} \bibinfo{person}{Shenghua Gao}.} \bibinfo{year}{2018}\natexlab{a}.
\newblock \showarticletitle{Gaze prediction in dynamic 360 immersive videos}. In \bibinfo{booktitle}{\emph{proceedings of the IEEE Conference on Computer Vision and Pattern Recognition}}. \bibinfo{pages}{5333--5342}.
\newblock


\bibitem[Yang et~al\mbox{.}(2019)]%
        {yang2019single}
\bibfield{author}{\bibinfo{person}{Qin Yang}, \bibinfo{person}{Junni Zou}, \bibinfo{person}{Kexin Tang}, \bibinfo{person}{Chenglin Li}, {and} \bibinfo{person}{Hongkai Xiong}.} \bibinfo{year}{2019}\natexlab{}.
\newblock \showarticletitle{Single and sequential viewports prediction for 360-degree video streaming}. In \bibinfo{booktitle}{\emph{2019 IEEE International Symposium on Circuits and Systems (ISCAS)}}. IEEE, \bibinfo{pages}{1--5}.
\newblock


\bibitem[Yaqoob et~al\mbox{.}(2020)]%
        {yaqoob2020survey}
\bibfield{author}{\bibinfo{person}{Abid Yaqoob}, \bibinfo{person}{Ting Bi}, {and} \bibinfo{person}{Gabriel-Miro Muntean}.} \bibinfo{year}{2020}\natexlab{}.
\newblock \showarticletitle{A survey on adaptive 360 video streaming: Solutions, challenges and opportunities}.
\newblock \bibinfo{journal}{\emph{IEEE Communications Surveys \& Tutorials}} \bibinfo{volume}{22}, \bibinfo{number}{4} (\bibinfo{year}{2020}), \bibinfo{pages}{2801--2838}.
\newblock


\bibitem[Yaqoob and Muntean(2021)]%
        {yaqoob2021combined}
\bibfield{author}{\bibinfo{person}{Abid Yaqoob} {and} \bibinfo{person}{Gabriel-Miro Muntean}.} \bibinfo{year}{2021}\natexlab{}.
\newblock \showarticletitle{A combined field-of-view prediction-assisted viewport adaptive delivery scheme for 360° videos}.
\newblock \bibinfo{journal}{\emph{IEEE Transactions on Broadcasting}} \bibinfo{volume}{67}, \bibinfo{number}{3} (\bibinfo{year}{2021}), \bibinfo{pages}{746--760}.
\newblock


\bibitem[Zhang et~al\mbox{.}(2022b)]%
        {zhang2022mfvp}
\bibfield{author}{\bibinfo{person}{Lei Zhang}, \bibinfo{person}{Weizhen Xu}, \bibinfo{person}{Donghuan Lu}, \bibinfo{person}{Laizhong Cui}, {and} \bibinfo{person}{Jiangchuan Liu}.} \bibinfo{year}{2022}\natexlab{b}.
\newblock \showarticletitle{MFVP: Mobile-Friendly Viewport Prediction for Live 360-Degree Video Streaming}. In \bibinfo{booktitle}{\emph{2022 IEEE International Conference on Multimedia and Expo (ICME)}}. IEEE, \bibinfo{pages}{1--6}.
\newblock


\bibitem[Zhang et~al\mbox{.}(2022a)]%
        {zhang2022vrformer}
\bibfield{author}{\bibinfo{person}{Zhihao Zhang}, \bibinfo{person}{Haipeng Du}, \bibinfo{person}{Shouqin Huang}, \bibinfo{person}{Weizhan Zhang}, {and} \bibinfo{person}{Qinghua Zheng}.} \bibinfo{year}{2022}\natexlab{a}.
\newblock \showarticletitle{VRFormer: 360-Degree Video Streaming with FoV Combined Prediction and Super resolution}. In \bibinfo{booktitle}{\emph{2022 IEEE Intl Conf on Parallel \& Distributed Processing with Applications, Big Data \& Cloud Computing, Sustainable Computing \& Communications, Social Computing \& Networking (ISPA/BDCloud/SocialCom/SustainCom)}}. IEEE, \bibinfo{pages}{531--538}.
\newblock


\bibitem[Zhu et~al\mbox{.}(2019)]%
        {zhu2019prediction}
\bibfield{author}{\bibinfo{person}{Yucheng Zhu}, \bibinfo{person}{Guangtao Zhai}, \bibinfo{person}{Xiongkuo Min}, {and} \bibinfo{person}{Jiantao Zhou}.} \bibinfo{year}{2019}\natexlab{}.
\newblock \showarticletitle{The prediction of saliency map for head and eye movements in 360 degree images}.
\newblock \bibinfo{journal}{\emph{IEEE Transactions on Multimedia}} \bibinfo{volume}{22}, \bibinfo{number}{9} (\bibinfo{year}{2019}), \bibinfo{pages}{2331--2344}.
\newblock


\end{thebibliography}










\end{document}